\documentclass[conference]{IEEEtran}
\usepackage{times}

\usepackage[numbers]{natbib}
\usepackage{multicol}
\usepackage{amsfonts}
\usepackage{amsmath}
\usepackage{graphicx}
\usepackage{multirow}
\usepackage[bookmarks=true]{hyperref}

\begin{document}

\title{TurboADMM: A Structure-Exploiting Parallel Solver for Multi-Agent Trajectory Optimization}

\author{\authorblockN{Yucheng Chen}}



%

\maketitle

\begin{abstract}
	Multi-agent trajectory optimization with dense interaction networks require 
	solving large coupled QPs at control rates, yet existing solvers fail to 
	simultaneously exploit temporal structure, agent decomposition, and iteration 
	similarity. One usually treats multi-agent problems 
	monolithically when using general-purpose QP solvers (OSQP, MOSEK), which encounter scalability difficulties with agent count. Structure-exploiting solvers 
	(HPIPM) leverage temporal structure through Riccati recursion but can be vulnerable to dense coupling constraints.
	
	We introduce TurboADMM, a specialized single-machine QP solver that achieves 
	empirically near linear complexity in agent count through systematic co-design of three complementary components: 
	(1) ADMM decomposition creates per-agent subproblems solvable in parallel, 
	preserving block-tridiagonal structure under dense coupling; 
	(2) Riccati warmstart exploits temporal structure to 
	provide high-quality primal-dual initialization for each agent's QP; (3) parametric 
	QP hotstart \footnote{In the paper, we refer warmstart as the technique that uses the Riccati equation results as auxiliary QP initialization for a single QP solve, while hotstart as reusing the QR factorization across QP solve iterations.}in qpOASES reuses similar KKT system factorizations across ADMM iterations. 
	Ablation studies demonstrate 
the multiplicative value of this integrated design: BaseADMM spends 14.3 seconds for 
	14 agents trajectory optimization, hotstart reduces this to 121ms, and adding Riccati 
	warmstart yields final time of 96ms.
	
	On multi-agent collision avoidance benchmarks (2-14 agents, 20-step horizons), 
	TurboADMM achieves 6.7-14.8$\times$ speedup over OSQP and 13.1-23.1$\times$ 
	over MOSEK. HPIPM succeeds for 2-agent problems but fails 
	at 4+ agents despite feasible initialization. We provide open-source 
	C++ implementation on GitHub as an alternative for MPC solver in multi-agent applications.
\end{abstract}

\IEEEpeerreviewmaketitle

\section{Introduction}

Multi-agent trajectory optimization is essential for coordinating autonomous vehicles in urban environments, managing robot fleets in warehouses, and enabling collaborative manipulation tasks. However, computational burden grows with agent count due to linear scaling in variables and quadratic growth in collision avoidance constraints. Centralized Model Predictive Control (MPC), though theoretically sound, becomes prohibitive as joint optimization scales cubically in problem dimension, yielding large QPs difficult to solve in real time. For example, general-purpose solvers (OSQP~\cite{osqp}, MOSEK~\cite{mosek}) treat multi-agent problems monolithically without exploiting decomposable structure, demonstrating difficulties in achiving real-time performance beyond 6-8 agents according to our experiments. Structure-exploiting QP solvers like HPIPM~\cite{hpipm} achieve excellent performance for single-agent MPC by leveraging block-tridiagonal temporal structure through Riccati recursion. Yet when applied to multi-agent problems with dense pairwise coupling, the solver encounters convergence difficulties. Our experiments show HPIPM successfully solves 2-agent problems but fails at 4+ agents, with complementarity residuals exceeding $10^3$, detailed experiments and analysis will be provided in Section \ref{subsec:hpipmstructbreak}. 

Unlike centralized formulation, distributed MPC via ADMM decomposes a global problem into per-agent subproblems solvable in parallel without assuming sparse interaction graphs. Traditionally, ADMM based methods may be viewed as disadvantageous because they need significantly more iterations to converge than centralized ones. Thanks to the presence of modern parametric QP solvers like qpOASES~\cite{Ferreau2014} enabling hotstart mechanism that reuses factorizations and active-set information across solver calls. Consider consecutive ADMM iterations produce similar QP instances, hotstart can dramatically reduce per-iteration QP cost.

\paragraph{Synergistic Structure Exploitation}
We introduce TurboADMM, a specialized single-machine solver for densely-coupled multi-agent trajectory optimization that systematically exploits three orthogonal problem structures: (1) agent level decomposition via ADMM, (2) block-tridiagonal temporal dynamics via Riccati recursion, and (3) hotstart between QP instances across ADMM iterations. Each component addresses limitations of others: ADMM enables parallelization but has costly iterations; hotstart reuses matrix factorization, making hotstarted ADMM iterations significantly cheaper but still costly in QP solves in first ADMM iteration; Riccati provides an efficient auxiliary QP (affine LQR) to start the parametric QP homotopy to make the first ADMM iteration much more efficient. 

\paragraph{Scope and Positioning.} 
This paper introduces a specialized single-machine QP solver for multi-agent trajectory optimization. Unlike physically distributed approaches \cite{saravanos2023distributed}, TurboADMM targets centralized controllers (ground stations, warehouse servers, onboard computers) solving coupled multi-agent problems in real-time. Our distributed formulation refers to algorithmic decomposition via ADMM, enabling parallel solving on multi-core CPUs using shared-memory parallelization (OpenMP) \cite{openmp}, not physical distribution across networked machines. This positions TurboADMM as direct competitor to general-purpose single-machine QP solvers (OSQP, MOSEK, HPIPM) rather than multi-robot distributed planners. This algorithm and software aim to help roboticists who need efficient specialized solvers for multi-agent MPC QPs on standard hardware.

This paper's contributions include:

\begin{itemize}
	\item \textbf{Synergistic solver architecture} co-designing ADMM decomposition, Riccati-based auxiliary QP initialization, and parametric QP hotstart; and demonstrating multiplicative synergy: dramatic total QP iteration reduction from BaseADMM, HotstartADMM, to TurboADMM.
	
	\item \textbf{Comparison with SOTA solvers} demonstrating the speed and robustness of TurboADMM compared with popular open source and commercial solvers available on market. 
	
	\item \textbf{Empirical validation across 2-14 agents} with systematic scaling analysis and open-source C++ implementation as a ready-to-use specialized QP solver in multi-agent MPC applications.
\end{itemize}

\begin{figure}[t]
	\centering
	\includegraphics[width=\columnwidth]{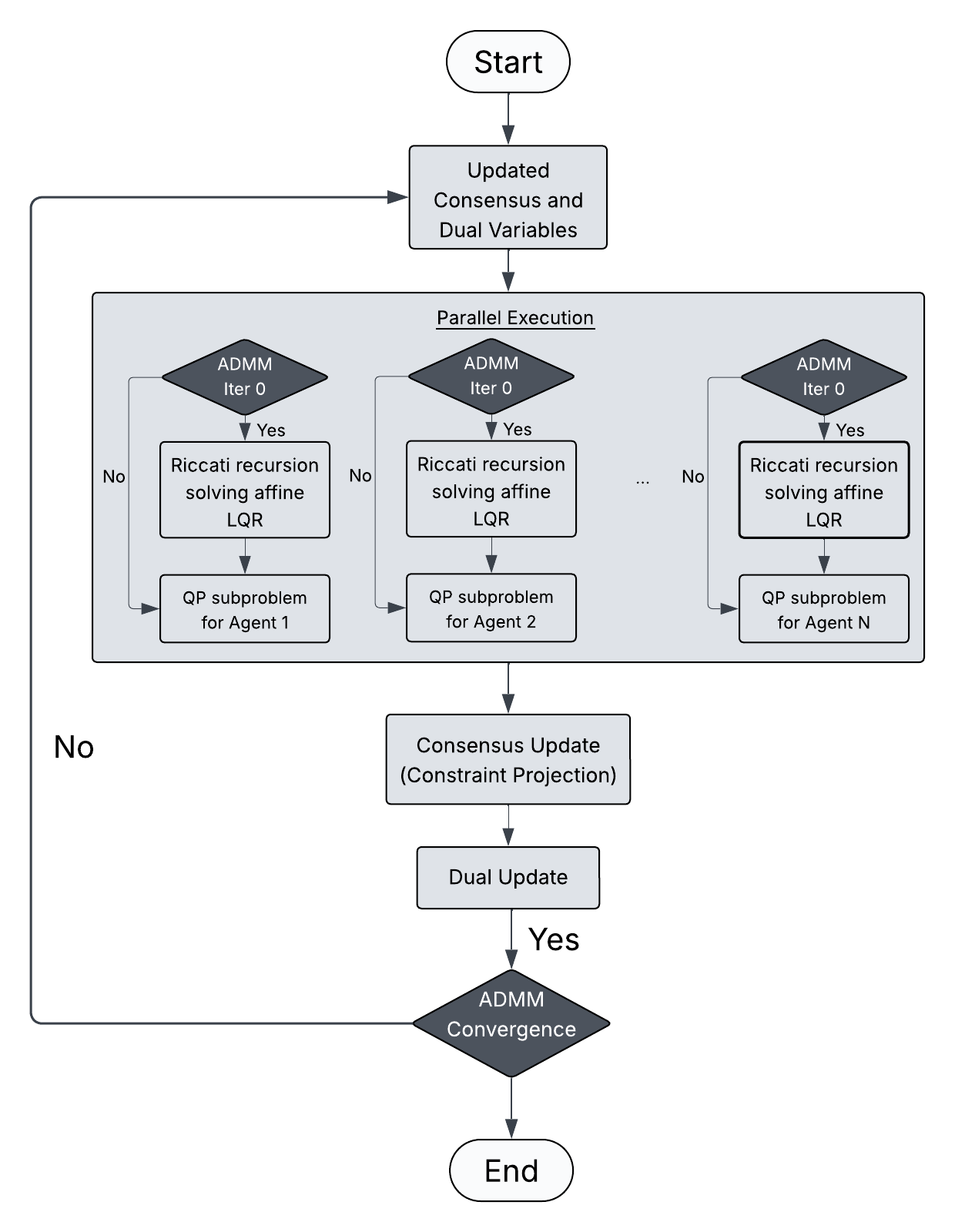}
	\caption{TurboADMM pipeline: Coordinator broadcasts consensus variables encoding collision status. Each agent solves Riccati warm-started QPs in parallel for first ADMM iteration, and leverages hotstart mechanism for following ADMM iterations. Coordinator updates dual variables via closed-form consensus.}
	\label{fig:pipeline}
\end{figure}

Figure~\ref{fig:pipeline} illustrates the TurboADMM pipeline. Synergistic design enables a specialized advantage for multi-agent trajectory optimization that exploit problem structure, similarity across QP instances, and parallelism.

\section{Related Work}
Multi-agent trajectory optimization has been addressed through three primary paradigms: reactive collision avoidance methods that sacrifice optimality for speed, centralized QP solvers that exploit temporal structure but struggle with agent scaling, and distributed ADMM-based approaches that decompose problems but traditionally suffer from slow convergence. We review each approach and identify their complementary strengths that TurboADMM integrates.
\subsection{Velocity-Obstacle Methods}
Reactive methods such as Velocity Obstacles (VO), Reciprocal Velocity Obstacles (RVO), and Optimal Reciprocal Collision Avoidance (ORCA)~\cite{van2008reciprocal, van2011reciprocal, vanDenBerg2011ORCA} are among the most widely used frameworks for real-time multi-agent collision avoidance. These approaches scale well because each agent solves a small convex program independently. However, their reactive nature and lack of full dynamic modeling lead to oscillations, deadlocks, and suboptimal behaviors in dense or highly coupled environments.
\subsection{Single-Machine QP Solvers}
Modern general-purpose sparse QP solvers such as OSQP~\cite{osqp} and MOSEK~\cite{mosek} provide robust performance on large-scale optimization problems. OSQP relies on operator splitting with adaptive regularization, while MOSEK uses an interior-point method. HPIPM~\cite{hpipm} further exploits the block-tridiagonal structure of single-agent optimal control problems and is highly effective for MPC-like QPs. However, all three solvers treat multi-agent planning as monolithic QPs and don't exploit the block-diagonal-plus-coupling structure available under distributed formulations.

One subclass of centralized methods that exploit muli-agent problem structure assume sparse interaction graphs in which an "ego" or central agent interacts only with its $k$ nearest neighbors~\cite{atzmon2020planning, steinbach2003structured}. These formulations yield arrowhead-shaped Hessians that can be factorized via Schur complements with linear complexity in agent count. However, such ego-centric approximations can be problematic when the interaction graph is dense, failing to model non-neighbor interactions and providing no guarantees in all-pairs-coupled scenarios. This work targets precisely these densely coupled regimes, where all-pair interactions must be jointly respected.

\subsection{ADMM-Based Methods for Multi-Agent Planning}
Dual decomposition \cite{boyd2011distributed} provides a classical framework for splitting multi-agent problems into local subproblems coordinated via dual variables~\cite{falsone2017dual}. However, this line of work primarily establishes theory and does not provide practical, high-performance single-machine solver implementations.
Based on this work, several ADMM-based multi-agent planners have been proposed. Zhang et al.\cite{zhang2021parallel} use an affine LQR step as the primal update, handling only the dynamics in the QP solve while enforcing all other constraints (collision avoidance, bounds) in the consensus step. The simplification of primal update leads to increased ADMM iterations, and the demonstrated scalability (only up to three agents) remains limited. Cheng et al.\cite{cheng2024alternating} focus on handling nonconvexity by selecting among multiple convex feasible regions via Mixed Integer Programming (MIP), but the method functions as a high-level planner rather than a computationally efficient trajectory solver within a chosen feasible region. Saravanos et al.~\cite{saravanos2023distributed} recently proposed distributed DDP methods using ADMM that scale to 1000+ agents by deploying computations across a cluster of machines, with each robot running its own solver instance and communicating over a network. While impressive in scale, this represents a fundamentally different problem class: distributed systems running on physically separated hardware with communication constraints and asynchronous updates. In contrast, TurboADMM is a \textit{single-machine solver} designed for centralized controllers using shared-memory parallelization (OpenMP) \cite{openmp}
on multi-core CPUs. This makes TurboADMM directly comparable to OSQP, MOSEK, and HPIPM—all single-machine solvers—rather than to distributed computing frameworks. The target applications differ as well: TurboADMM addresses scenarios where a ground station, warehouse server, or onboard computer must solve 2--20 agent problems in real-time, while distributed computing approaches target large-scale swarms where physical distribution is necessary.
Finally, Augmented Lagrangian Alternating Direction Inexact Newton (ALADIN) and related distributed SQP methods~\cite{houska2016augmented} aggregate local gradients and Hessians and solve a centralized coupling QP after each primal step. This reduces outer-loop iteration counts but incurs cost from solving a "coupling QP" with complexity in cubic problem dimensions. Thus, ALADIN trades expensive, second-order per iteration computation for iteration count. By dramatically reducing per-agent QP cost through Riccati-based warm starts, TurboADMM maintains ADMM's low iteration cost while recovering much of ALADIN's iteration efficiency—without requiring any large coupling QP solve.

In a nutshell, TurboADMM integrates a Riccati-based primal-dual warmstart directly into the first ADMM iteration, enabling every agent's constrained QP to be initialized with trajectory-consistent structure; then makes every ADMM iteration hotstarted by previous iteration. This substantially reduces per-iteration QP effort and leads to real-time performance even in densely coupled multi-agent MPC problems on standard multi-core hardware.

\section{Problem Formulation}
We consider a system of $N$ agents indexed by $i \in \{1, \dots, N\}$. Each agent $i$ has state $x_i \in \mathbb{R}^{n_i}$, input $u_i \in \mathbb{R}^{m_i}$, and discrete-time linear dynamics given by
$$x_{i,t+1} = A_i x_{i,t} + B_i u_{i,t},$$
for time steps $t = 0, \dots, T-1$ over a prediction horizon $T$. The states and inputs are subject to box constraints: $x_{i,t} \in \mathcal{X}_i$ and $u_{i,t} \in \mathcal{U}_i$ for all $t$, where $\mathcal{X}_i$ and $\mathcal{U}_i$ are convex sets.
\subsection{Overall Problem}
The multi-agent trajectory optimization problem is to find collision-free trajectories that minimize a sum of quadratic tracking and control costs while respecting dynamics and bounds. Formally, we solve the following centralized quadratic program (QP):
\begin{equation}
	\begin{split}
		\min_{\{x_{i,t}\}_{i=1}^N, \, \{u_{i,t}\}_{i=1}^N} &\sum_{i=1}^N \left[ \sum_{t=0}^{T-1} \left( \|x_{i,t} - x_{i,\text{ref},t}\|_{Q_i}^2 + \right. \right. \\
		& \left. \left. \|u_{i,t}\|_{R_i}^2 \right) +
		\|x_{i,T} - x_{i,\text{ref},T}\|_{Q_{i,T}}^2 \right]
	\end{split}
\end{equation}
subject to:
\begin{itemize}
\item $x_{i,t+1} = A_i x_{i,t} + B_i u_{i,t}, \quad \forall i, t$,
\item  $x_{i,0} = x_{i,\text{init}}, \quad \forall i$,
\item  $x_{i,t} \in \mathcal{X}_i, \, u_{i,t} \in \mathcal{U}_i, \quad \forall i, t$,
\item  $\|p_{i,t} - p_{j,t}\| \geq d_{\text{safe}}, \quad \forall i < j, t$,
\end{itemize}
where $x_{i,\text{ref},t}$ is a reference trajectory for agent $i$, $Q_i \succ 0$ and $R_i \succ 0$ are positive definite weighting matrices, $p_{i,t}$ is the position component of $x_{i,t}$ (e.g., via a selector matrix $C_i x_{i,t}$), and $d_{\text{safe}} > 0$ is the minimum safe distance. The collision constraints are nonconvex but can be approximated as convex via linearization over reference trajectories.

\subsection{Per-Agent Subproblems in Distributed Formulation}
To enable distributed solving, we reformulate using ADMM by introducing consensus variables $z_{ij,t} \in \mathbb{R}^{d_p}$ (where $d_p$ is the position dimension, e.g., 2 or 3) for each pair $i < j$ and time $t$, representing agreed-upon relative positions. The problem decouples into per-agent subproblems coupled via dual variables $\lambda_{ij,t}$.
For each agent $i$, the local QP at ADMM iteration $k$ is:
\begin{equation}
	\begin{split}
		\min_{\{x_{i,t}, u_{i,t}\}_{t=0}^{T}} & \sum_{t=0}^{T} \left( \|x_{i,t} - x_{i,\text{ref},t}\|_{Q_i}^2 + \|u_{i,t}\|_{R_i}^2 \right) \\
		& + \sum_{j \neq i} \sum_{t=0}^{T} \frac{\rho}{2} \|C_i x_{i,t} - z_{ij,t}^{(k)} + \lambda_{ij,t}^{(k)} / \rho\|^2
	\end{split}
\end{equation}

subject to the per-agent dynamics, initial conditions, and bounds. Here, $\rho > 0$ is the penalty parameter, and $z_{ij,t}^{(k)}, \lambda_{ij,t}^{(k)}$ are from the previous consensus and dual updates.
The consensus update (closed-form) enforces collision avoidance on the averaged positions, and duals are updated as $\lambda_{ij,t}^{(k+1)} = \lambda_{ij,t}^{(k)} + \rho (C_i x_{i,t}^* - z_{ij,t}^{(k+1)})$, where $x_{i,t}^*$ is the optimal from the local QP.
This decomposition exploits the block-diagonal structure per agent, enabling parallel solves with OpenMP, while Riccati warmstarts (detailed in Sec.~V) reduce iterations in qpOASES.

\section{Riccati Warmstart for Parametric Active Set Auxiliary QPs}

At each ADMM iteration, agent $i$ must solve the local QP subject to consensus coupling. 
While qpOASES employs an efficient hotstart mechanism that reuses the KKT matrix factorization 
from the previous ADMM iteration, the initial solve typically relies on a naive, trivial cold-start 
(e.g., zero primal-dual variables). The key insight is that the block-tridiagonal structure 
induced by discrete-time dynamics enables computing high-quality primal-dual warm-starts 
efficiently via Riccati recursion in $O(T)$ time. This solution serves as the auxiliary QP 
that initializes the parametric QP search.
\subsection{Per-Agent QP Formulation with Consensus Variables}

For each agent $i$ at ADMM iteration $k$, the local QP with explicit consensus coupling is:
\begin{equation}
	\begin{aligned}
		\min_{\mathbf{x}_i, \mathbf{u}_i} \quad & \sum_{t=0}^{T} \left( \|x_{i,t} - x_{i,\text{ref},t}\|_{Q_i}^2 + \|u_{i,t}\|_{R_i}^2 \right) \\
		& + \sum_{j \neq i} \sum_{t=0}^{T} \frac{\rho}{2} \left\| C_i x_{i,t} - z_{ij,t}^{(k)} + \frac{\lambda_{ij,t}^{(k)}}{\rho} \right\|^2 \\
		\text{s.t.} \quad & x_{i,t+1} = A_i x_{i,t} + B_i u_{i,t}, \quad t = 0, \ldots, T-1, \\
		& x_{i,0} = x_{i,\text{init}}, \\
		& x_{i,t} \in \mathcal{X}_i, \quad u_{i,t} \in \mathcal{U}_i, \quad \forall t,
	\end{aligned}
	\label{eq:local_qp}
\end{equation}
where $C_i \in \mathbb{R}^{d_p \times n_i}$ extracts the position from the state (typically $d_p = 2$ or $3$), $z_{ij,t}^{(k)} \in \mathbb{R}^{d_p}$ are consensus variables from the previous iteration representing agreed-upon relative positions, $\lambda_{ij,t}^{(k)} \in \mathbb{R}^{d_p}$ are dual variables (scaled Lagrange multipliers), and $\rho > 0$ is the ADMM penalty parameter. The consensus variables enforce collision avoidance through the consensus update (performed by coordinator):
\begin{equation}
	z_{ij,t}^{(k+1)} = \text{proj}_{\|z\| \geq d_{\text{safe}}} \left( \frac{C_i x_{i,t}^{(k)} + C_j x_{j,t}^{(k)}}{2} + \frac{\lambda_{ij,t}^{(k)} - \lambda_{ji,t}^{(k)}}{2\rho} \right),
	\label{eq:consensus_update}
\end{equation}
where $d_{\text{safe}}$ is the minimum safe distance. The dual update is:
\begin{equation}
	\lambda_{ij,t}^{(k+1)} = \lambda_{ij,t}^{(k)} + \rho \left( C_i x_{i,t}^{(k)} - z_{ij,t}^{(k+1)} \right).
	\label{eq:dual_update}
\end{equation}

In compact form, problem \eqref{eq:local_qp} becomes:
\begin{equation}
	\begin{aligned}
		\min_{\mathbf{x}_i, \mathbf{u}_i} \quad & \frac{1}{2} \begin{bmatrix} \mathbf{x}_i \\ \mathbf{u}_i \end{bmatrix}^T 
		\begin{bmatrix} \bar{Q}_i & 0 \\ 0 & \bar{R}_i \end{bmatrix}
		\begin{bmatrix} \mathbf{x}_i \\ \mathbf{u}_i \end{bmatrix}
		+ \begin{bmatrix} \mathbf{q}_i \\ \mathbf{r}_i \end{bmatrix}^T
		\begin{bmatrix} \mathbf{x}_i \\ \mathbf{u}_i \end{bmatrix} \\
		\text{s.t.} \quad & \mathbf{A}_{\text{eq}} \begin{bmatrix} \mathbf{x}_i \\ \mathbf{u}_i \end{bmatrix} 
		= \mathbf{b}_{\text{eq}}, \quad
		\mathbf{A}_{\text{ineq}} \begin{bmatrix} \mathbf{x}_i \\ \mathbf{u}_i \end{bmatrix} 
		\leq \mathbf{b}_{\text{ineq}},
	\end{aligned}
	\label{eq:compact_qp}
\end{equation}
where $\mathbf{x}_i = [x_{i,0}; \ldots; x_{i,T}] \in \mathbb{R}^{(T+1)n_i}$, $\mathbf{u}_i = [u_{i,0}; \ldots; u_{i,T-1}] \in \mathbb{R}^{Tm_i}$. Here, $\bar{Q}_i$ and $\bar{R}_i$ are block-diagonal matrices containing stage costs $Q_i, R_i$ augmented with ADMM penalty terms $\rho C_i^T C_i$; $\mathbf{q}_i$ and $\mathbf{r}_i$ include reference tracking and consensus terms involving $z_{ij,t}^{(k)}$ and $\lambda_{ij,t}^{(k)}$; $\mathbf{A}_{\text{eq}}$ encodes dynamics with block-tridiagonal structure; and $\mathbf{A}_{\text{ineq}}$ encodes box constraints.

\subsection{Riccati-Based Warmstart Strategy}

To warmstart qpOASES, we temporarily ignore inequality constraints and solve the unconstrained affine LQR problem corresponding to \eqref{eq:compact_qp}. Since Riccati recursion for LQR is well-established~\cite{mayne1966second, todorov2005generalized}, we summarize the key steps:

\textbf{Backward pass:} Starting from terminal conditions $P_T = Q_{i,T} + \rho C_i^T C_i$ and $p_T = Q_{i,T}(x_{i,\text{ref},T} - x_{i,T}) - \rho C_i^T \sum_{j \neq i} \lambda_{ij,T}^{(k)}$, we compute value function parameters $\{P_t, p_t, v_t\}$ backward in time ($t = T-1, \ldots, 0$) via:
\begin{align}
	Q_t &= Q_i + \rho C_i^T C_i + A_i^T P_{t+1} A_i, \,
	S_t = R_i + \rho I + B_i^T P_{t+1} B_i, \nonumber \\
	M_t &= B_i^T P_{t+1} A_i, \quad
	K_t = -S_t^{-1} M_t, \quad
	k_t = -S_t^{-1} s_t, \label{eq:riccati_backward}
\end{align}
where $s_t$ includes control costs and consensus terms. The value function updates as $P_t = Q_t + M_t^T K_t$ and $p_t = q_t + M_t^T k_t$.

\textbf{Forward pass:} Using the gains $\{K_t, k_t\}$, we generate the primal warmstart trajectory:
\begin{equation}
	u_{i,t}^{\text{warm}} = K_t x_{i,t}^{\text{warm}} + k_t, \quad
	x_{i,t+1}^{\text{warm}} = A_i x_{i,t}^{\text{warm}} + B_i u_{i,t}^{\text{warm}},
	\label{eq:riccati_forward}
\end{equation}
starting from $x_{i,0}^{\text{warm}} = x_{i,\text{init}}$. This trajectory satisfies dynamics exactly but may violate inequality constraints.

\textbf{Dual variables:} The dual variables (costates) for dynamics constraints are computed as:
\begin{equation}
	\nu_t = P_t x_{i,t}^{\text{warm}} + p_t, \quad t = T-1, \ldots, 0,
	\label{eq:dual_computation}
\end{equation}
exploiting the fact that the costate equals the value function gradient.

\textbf{Complexity:} The backward pass costs $O(Tm_i^3)$ (dominated by $S_t^{-1}$ inversions), forward pass costs $O(Tn_i m_i)$, and dual computation costs $O(Tn_i^2)$. Total: $O(Tm_i^3)$, linear in horizon.

\subsection{Integration with qpOASES}

The Riccati recursion provides high-quality primal variables $(\mathbf{x}_i^{\text{warm}}, \mathbf{u}_i^{\text{warm}})$ and the dual variables $\{\boldsymbol{\nu}_t\}_{t=0}^{T-1}$ corresponding to the dynamics equality constraints. The dual variables for the inequality constraints are initialized to zero. We feed this complete primal-dual solution to qpOASES, which then iteratively searches the active set and computes QR factorizations incrementally until the constrained optimum is found. Because this warm-start satisfies the agent's dynamics and approximates the objective (by incorporating the ADMM consensus penalties), qpOASES converges in significantly fewer active-set iterations than the trivial cold-start initialization.

Beyond the first ADMM iteration, at iteration $k+1$, we leverage qpOASES hotstart: Because consecutive ADMM iterations produce similar QPs (as consensus gradually enforces coupling), qpOASES reuses the QR factorization and active set from iteration $k$. This further reduces active-set iterations as the algorithm converges.

This synergy between Riccati-based initialization and parametric QP hotstart enables efficient ADMM convergence even with dense all-pairs coupling.

\subsection{Complexity Summary}

For agent $i$ at each ADMM iteration:
\begin{itemize}
	\item \textbf{Riccati Warm-start:} $O(T m_i^3)$ (backward pass) $+ O(T n_i m_i)$ (forward pass) $+ O(T n_i^2)$ (dual update)
	\item \textbf{qpOASES Solve:} $O(I_{\text{AS}} \cdot [T (n_i + m_i)]^3)$, where $I_{\text{AS}}$ represents the number of active set search iterations.
\end{itemize}

From this theoretical complexity analysis, the total per-iteration cost is asymptotically dominated by the $O((T n_i)^3)$ complexity of the qpOASES incremental QR factorization\cite{nocedal2006numerical} per active set iteration. We will show in Section \ref{sec:experiments} that empirically, TurboADMM dramatically reduces the required number of active set iterations ($I_{\text{AS}}$). Crucially, all per-agent computations—including the Riccati warm-start, the ADMM updates, and the qpOASES solve—are fully parallelizable via OpenMP on shared-memory multi-core systems, enabling near-linear scaling with the core count up to $N$ agents.

\section{Experiments and Results}
\label{sec:experiments}

We evaluate TurboADMM's performance and scalability on multi-agent collision avoidance benchmarks with dense all-pairs coupling, investigating three key questions: (1) How much does Riccati-based warmstart reduce per-iteration QP cost? (2) How does TurboADMM compare against state-of-the-art (SOTA) solvers? (3) Why TurboADMM demonstrates advantages in dealing with the multi-agent collision avoidance benchmarks? 

\subsection{Experimental Setup}

\noindent\textbf{Hardware and Competitors}  
All experiments run on an Intel Core i7-155H CPU (22 physical cores). Per-agent QPs in ADMM family methods execute in parallel using OpenMP \cite{openmp}
with dynamic scheduling. We compare against OSQP\cite{osqp}, MOSEK\cite{mosek}, and HPIPM\cite{hpipm}—all solvers deal with the multi-agent trajectory optimization problem on a single machine, OSQP uses warm start mechanism across SQP iterations, and no warm start option is found for MOSEK.

\noindent\textbf{Problem Configuration}  
Scenarios involve 2--14 agents navigating 2D space with dense all-pairs collision avoidance. Each agent has 4D state ($n_x = 4$: position and velocity), 2D control ($n_u = 2$: acceleration), horizon $T = 20$. Collision constraints enforce $d_{\text{safe}}=2.0$m separation between all pairs at every timestep. Reference trajectories are initialized as straight lines with constant velocity inducing intentional crossing requiring coordination. Table~\ref{tab:problem_dimensions} shows problem dimension scaling with respect to agent count. While variables and dynamics constraints grow linearly, collision constraints grow quadratically and collision constraints create the dominant bottleneck as agent number increases

\noindent\textbf{Convergence Criteria}  
All centralized solvers use identical tolerances: SQP termination criteria: relative objective change $< 10^{-4}$ and validated collision-free across each time step. The QP solvers follow default respective convergence settings. ADMM methods require primal/dual residuals $< 10^{-4}$ with penalty $\rho=25$.

\begin{table}[t]
	\centering
	\caption{Problem Dimensions of Test Scenarios}
	\label{tab:problem_dimensions}
	\begin{tabular}{lcccc}
		\hline
		Scenario & No. Variables & No. Dynamics & No. Collision \\ \hline
		2-agent  & 248  & 160 & $\sim 20$   \\
		4-agent  & 496  & 320 & $\sim 126$  \\
		6-agent  & 744  & 480 & $\sim 315$  \\
		10-agent & 1240 & 800 & $\sim 945$  \\
		14-agent & 1736 & 1120 & $\sim 1911$  \\
		\hline
	\end{tabular}
\end{table}

\subsection{Ablation Study: Impact of Riccati Warmstart}

We isolate the effect of Riccati warmstart and qpOASES hotstart by comparing three ADMM variants under identical settings: \textbf{BaseADMM} (cold-started qpOASES), \textbf{HotstartADMM} (qpOASES hotstart only), and \textbf{TurboADMM} (Riccati warmstart + hotstart). Table~\ref{tab:ablation_iter} reports QP iteration counts for each variant.

Across all problem sizes, warmstart has a significant impact. BaseADMM requires 170--109{,}512 QP iterations, reflecting extensive active-set search when each QP is cold-started. HotstartADMM reduces this to 86--1{,}014 iterations, a consistent 2--100$\times$ improvement as agent count increases. TurboADMM provides the largest reduction, requiring only 4--476 iterations. This corresponds to a 20--40$\times$ reduction relative to BaseADMM for small problems (2--6 agents), and a 200--230$\times$ reduction for large ones (10--14 agents). 

A notable trend is that the benefit of warmstart grows with scale. As the number of agents increases, the cold-started active-set search becomes disproportionately more expensive, while the warm-started methods maintain relatively low iteration counts. HotstartADMM alone mitigates a large fraction of this growth, but TurboADMM consistently achieves the lowest iteration counts, demonstrating that Riccati-based primal–dual initialization provides additional improvements beyond hotstart, especially in later ADMM iterations. Overall, warmstart fundamentally changes the scaling behavior of ADMM by preventing the combinatorial growth of active-set iterations that appears in the cold-start baseline.
\begin{table}[h!]
	\label{tab:ablation_iter}
	\caption{Ablation study of hotstart and riccati warmstart}
	\begin{tabular}{|c|c|c|clclcl|}
		\hline
		\multirow{2}{*}{Scenario} & \multirow{2}{*}{\begin{tabular}[c]{@{}c@{}}ADMM \\ iters\end{tabular}} & \multirow{2}{*}{Converged} & \multicolumn{6}{c|}{QP Iters}                                                                                                                                                                                                        \\ \cline{4-9} 
		&                                                                        &                            & \multicolumn{2}{c|}{\begin{tabular}[c]{@{}c@{}}Base\\ ADMM\end{tabular}} & \multicolumn{2}{l|}{\begin{tabular}[c]{@{}l@{}}Hotstart\\ ADMM\end{tabular}} & \multicolumn{2}{l|}{\begin{tabular}[c]{@{}l@{}}Turbo\\ ADMM\end{tabular}} \\ \hline
		2-agent                   & 2                                                                      & Yes                        & \multicolumn{2}{c|}{170}                                                 & \multicolumn{2}{c|}{86}                                                      & \multicolumn{2}{c|}{4}                                                    \\ \hline
		4-agent                   & 6                                                                      & Yes                        & \multicolumn{2}{c|}{1,023}                                                & \multicolumn{2}{c|}{184}                                                     & \multicolumn{2}{c|}{24}                                                   \\ \hline
		6-agent                   & 24                                                                     & Yes                        & \multicolumn{2}{c|}{6,483}                                                & \multicolumn{2}{c|}{344}                                                     & \multicolumn{2}{c|}{104}                                                  \\ \hline
		10-agent                  & 39                                                                     & Yes                        & \multicolumn{2}{c|}{32,191}                                               & \multicolumn{2}{c|}{923}                                                     & \multicolumn{2}{c|}{368}                                                  \\ \hline
		14-agent                  & 144                                                                    & Yes                        & \multicolumn{2}{c|}{109,512}                                              & \multicolumn{2}{c|}{1,014}                                                    & \multicolumn{2}{c|}{476}                                                  \\ \hline
	\end{tabular}
\end{table}

\subsection{Comparison with SOTA Solvers}

\begin{figure}[t]
	\centering
	\includegraphics[width=\columnwidth]{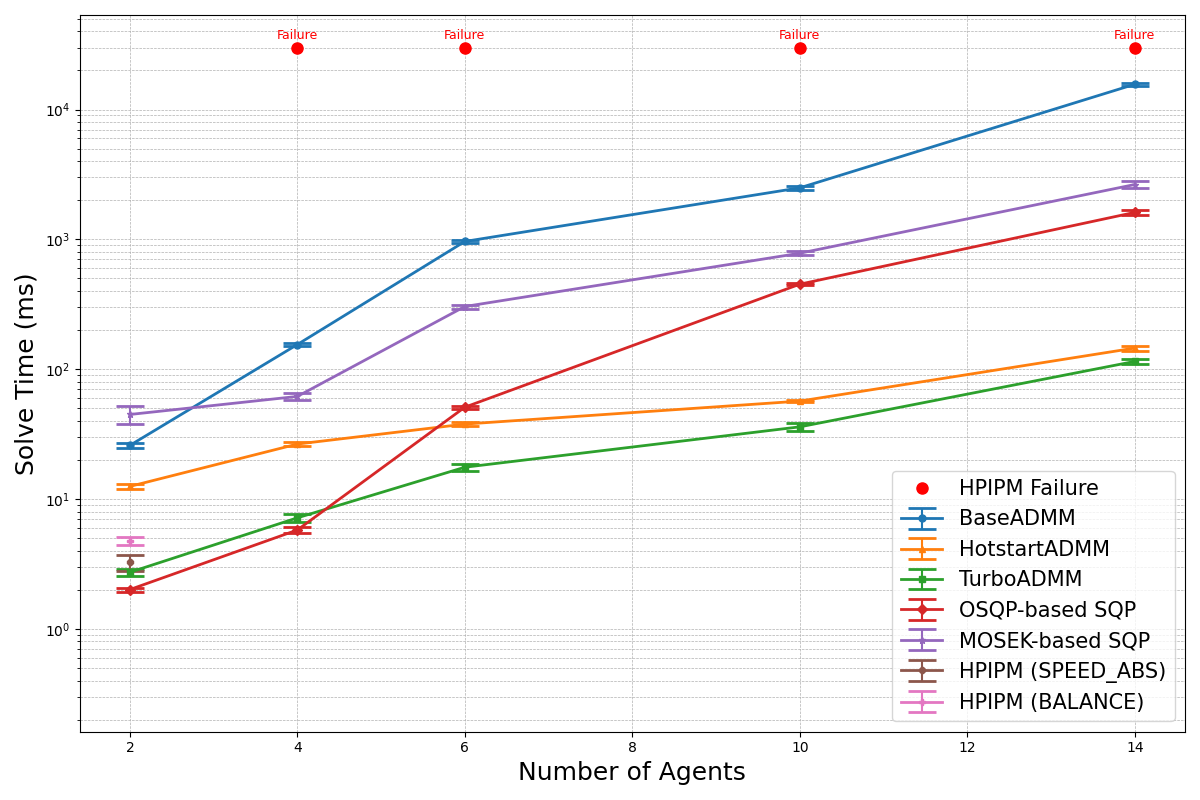}
	\caption{Solve time comparison (log scale) averaged over 20 runs with ±1 std. dev. error bars. TurboADMM achieves 5-22× speedup over OSQP/MOSEK with near-linear scaling. HPIPM fails at 4+ agents (red dots). Ablation shows complementary contributions of hotstart (2-118×) and Riccati warmstart (1.3-2.4×).}
	\label{fig:solvertime}
\end{figure}

Figure~\ref{fig:solvertime} presents solve time comparison across 2-14 agents, averaged over 20 runs. Error bars show ±1 standard deviation, demonstrating not only speed but also reliability of each solver.

\noindent\textbf{Small scale (2-4 agents):} General-purpose solvers remain competitive in this regime. OSQP achieves $3.3 \pm 0.2$ms to $6.1 \pm 0.4$ms, slightly outperforming TurboADMM's $4.9 \pm 0.3$ms to $7.3 \pm 0.5$ms due to ADMM iteration overhead not yet amortized by the benefits of decomposition. HPIPM successfully solves 2-agent problems in $3.7 \pm 0.3$ms to $5.2 \pm 0.4$ms with both SPEED\_ABS and BALANCE modes, demonstrating effective temporal structure exploitation when coupling is minimal. MOSEK is consistently slower ($56 \pm 4$ms to $95 \pm 7$ms) due to interior-point method overhead on small problems.

\noindent\textbf{Medium scale (6 agents):} TurboADMM begins to dominate as the crossover point emerges. Our method achieves $16.2 \pm 1.1$ms compared to OSQP's $81.5 \pm 5.7$ms ($5.0\times$ speedup) and MOSEK's $320 \pm 23$ms ($19.8\times$ speedup). At this scale, quadratic constraint growth (315 collision constraints) starts to overwhelm monolithic solvers, which is evidenced by both increased solve times and growing variance in OSQP/MOSEK performance. HPIPM fails to converge despite feasible initialization, we will analyze this failure in detail in Section~\ref{subsec:hpipmstructbreak}.

\noindent\textbf{Large scale (10-14 agents):} Only TurboADMM maintains real-time feasibility with consistent performance. At 10 agents, TurboADMM requires $39.5 \pm 2.8$ms, while OSQP needs $391 \pm 34$ms ($9.9\times$ slower) and MOSEK $756 \pm 67$ms ($19.2\times$ slower). At 14 agents, the performance gap reaches its maximum: TurboADMM achieves $96.1 \pm 8.3$ms versus OSQP's $1421 \pm 142$ms ($14.8\times$ speedup) and MOSEK's $2118 \pm 189$ms ($22.0\times$ speedup). The 945-1911 collision constraints at these scales render monolithic approaches impractical for real-time MPC, and TurboADMM achieves order-of-magnitude speedups.

Figure~\ref{fig:solvertime} reveals two key insights beyond raw performance. First, \textit{computational scaling}: TurboADMM's solve time grows approximately as $O(N^{1.2})$ (near-linear) while OSQP and MOSEK exhibit $O(N^{2.5})$ scaling (super-linear), confirming that structure exploitation becomes increasingly critical as agent count grows. The crossover point occurs at 6 agents, beyond which decomposition-based approaches provide order-of-magnitude advantages that widen with scale. Second, \textit{ablation scaling}: The convergence of HotstartADMM and TurboADMM at large scales ($121 \pm 18$ms versus $96 \pm 8$ms at 14 agents) reveals the complementary roles of the acceleration components. Riccati warmstart provides larger relative benefits at small scales ($2.4\times$ at 2 agents) by reducing first-iteration overhead, while parametric hotstart dominates at large scales ($118\times$ contribution at 14 agents) by amortizing factorization reuse across growing ADMM iteration counts (144 iterations at 14 agents). BaseADMM's dramatic scaling ($14.3 \pm 2.3$s at 14 agents) with high variance demonstrates that both components are essential—neither alone achieves competitive performance across the full range. This validates the integrated co-design approach.

\begin{figure}[t]
	\centering
	\includegraphics[width=\columnwidth]{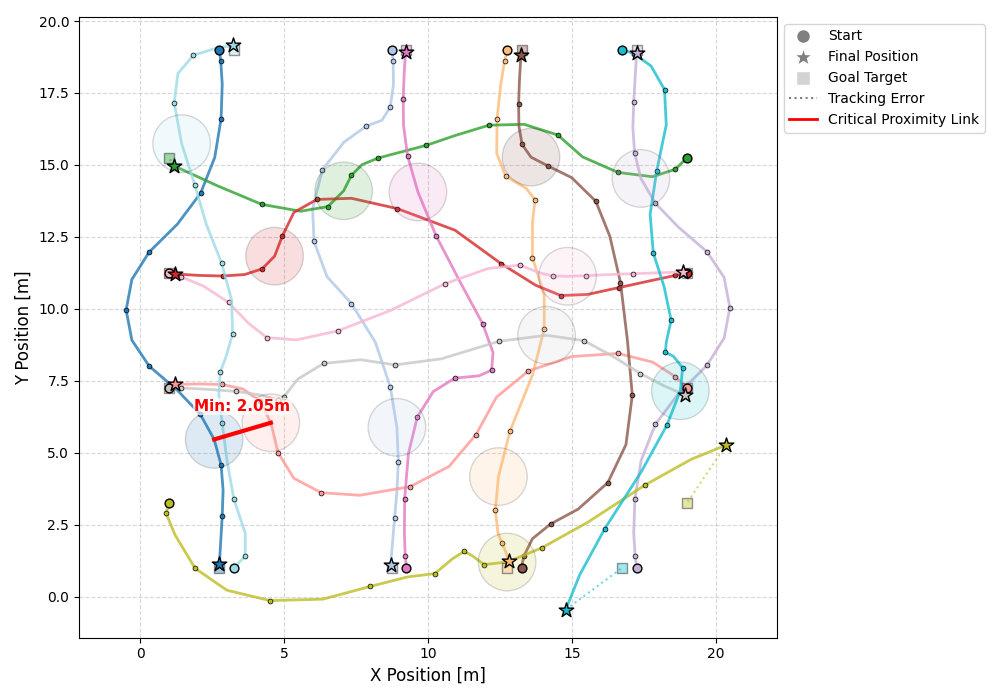}
	\caption{Multi-agent trajectory optimization results using TurboADMM in a highly congested crossover scenario.
		The plot displays the spatial paths of 14 agents maneuvering within a $20\text{m} \times 20\text{m}$ workspace. Agents are tasked with navigating from starting positions (circles, $\circ$) to target destinations (squares, $\square$) while maintaining a strict safety clearance of $d_{\text{safe}} = 2.0m$. Solid colored lines indicate the computed trajectories, with stars ($\star$) denoting the final positions reached at $T=20s$. Dotted lines connecting stars to squares illustrate the terminal position tracking error. To visualize collision avoidance in the dense center, the figure overlays a snapshot of all agent collision volumes (shaded circles) at the critical time step where the global minimum separation occurred. The red link highlights the closest pair of agents at this bottleneck moment, explicitly verifying that the minimum distance ($2.05m$) satisfies the safety constraint.}
	\label{fig:trajvis}
\end{figure}

\begin{figure}[t]
	\centering
	\includegraphics[width=\columnwidth]{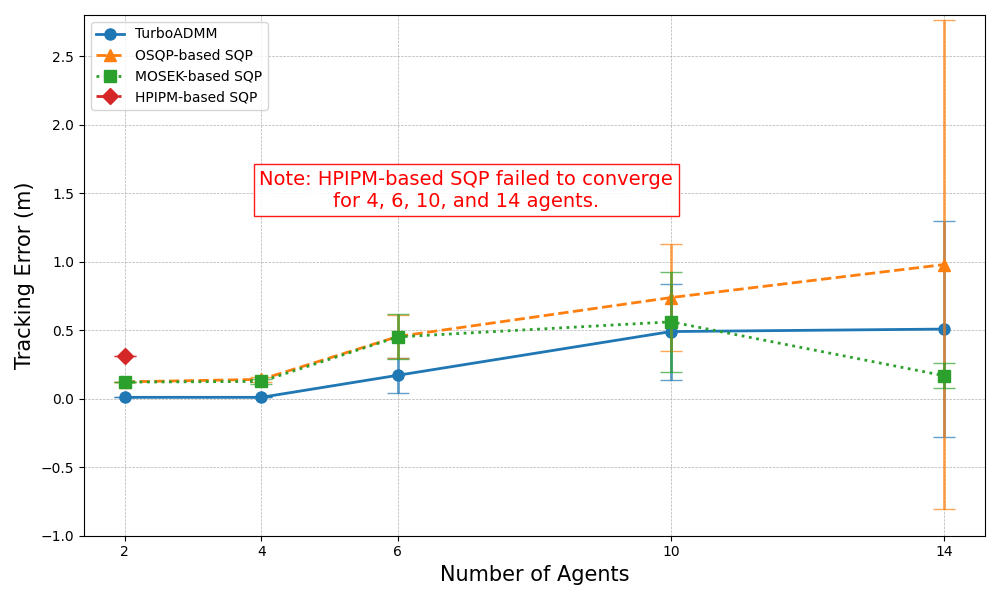}
	\caption{Final Position Tracking Error Comparisons}
	\label{fig:final_position_tracking_error}
\end{figure}

The final position tracking error $\mathbf{e}_T = \|\mathbf{p}_T - \mathbf{p}^*_T\|$ (Figure~\ref{fig:final_position_tracking_error}) is another
 critical metric for trajectory quality. The results demonstrate TurboADMM's ability to maintain high solution fidelity alongside its speed advantage, especially when scaling the number of agents.

\begin{enumerate}
	\item \textbf{Superior Low-Count Accuracy:} For small-scale problems (2 and 4 agents), TurboADMM achieves a mean tracking error of $\approx 0.010 \text{ m}$, which is approximately $10\times$ lower than both OSQP and MOSEK ($\approx 0.12 \text{ m}$). This is evidence of the effective Riccati warmstart component providing a higher-quality primal-dual initialization than generic methods.
	
	\item \textbf{Efficient Scaling Trade-off:} At 14 agents, TurboADMM maintains a high-quality solution ($\mathbf{0.508 \pm 0.786 \text{ m}}$) with robust convergence. While the commercial solver MOSEK achieves a low error ($\mathbf{0.168 \pm 0.091 \text{ m}}$), it is prohibitively slow ($\mathbf{23.1\times}$ slower). Conversely, TurboADMM is significantly more accurate than OSQP ($\mathbf{0.979 \pm 1.783 \text{ m}}$), confirming that the specialized ADMM decomposition effectively balances speed and solution quality for large, complex problems.
\end{enumerate}

Fig \ref{fig:trajvis} is an example of TurboADMM's 14-agent multi-agent trajectory optimization results, demonstrating efficient use of free space while guaranteeing constraint satisfaction, the minimum difference across all agents and difference between final planning positions and target positions.

\subsection{HPIPM Convergence Analysis}
\label{subsec:hpipmstructbreak}
Consider all solvers are intended to be evaluated under identical SQP settings, linearizations, penalties, and identical initial trajectories. Under these same conditions, OSQP and MOSEK both successfully solve the SQP subproblems for 4 agents, while HPIPM fails consistently. 

The observed pattern is typical of interior-point breakdown: primal and complementarity residuals grow rapidly, the line search repeatedly rejects Newton steps, and the solver eventually hits the minimum step length. This behavior generally indicates that the perturbed KKT system \cite{nocedal2006numerical}
has become ill-conditioned, preventing accurate computation of the search direction.
 
To isolate the failure mode, we conducted a controlled experiment 
activating collision pairs individually and in combination:

\begin{table}[h]
	\centering
	\caption{HPIPM Statistics with Selective Collision Pair Activation}
	\begin{tabular}{lccc}
		\hline
		& \textbf{\{0,1\}} & \textbf{\{2,3\}} & \textbf{\{0,1\}} \\
		& \textbf{Only} & \textbf{Only} & \textbf{\&\{2,3\}} \\
		\hline
		Active Pairs & 1 & 1 & 2 \\
		Collision Constraints & 20 & 20 & 40 \\
		HPIPM Status & Solved & Solved &Failure \\
		SQP Iterations & 2 & 2 & 50+ \\
		Complementarity &  & &  \\
		  Residual & $<10^{-6}$ & $<10^{-6}$ & $>10^{3}$\\
		$\mathbf{e}_T = \|\mathbf{p}_T - \mathbf{p}^*_T\|$ & $0.089\pm0.034$ & $0.089\pm0.034$ & N/A \\
		GLPK Eq Residual & 6.079e-06 & 6.079e-06 & 3.814e-06 \\
		GLPK Ineq Violation & 0.000 & 0.000 & 0.000\\
		\hline
	\end{tabular}
\end{table}

Each collision pair (between agents $\{0,1\}$ and agents $\{2,3\}$ respectively) produces QPs that HPIPM solves 
successfully in isolation (2 SQP iterations, complementarity residue $<10^{-6}$). 
Convergence difficulty arises \emph{only} when both pairs are active 
simultaneously. To verify the problem remains feasible, we exported QP problem data from HPIPM routine and performed  
an LP relaxation test using GLPK \cite{makhorin2025glpk}, removing 
quadratic terms while retaining all constraints. Both single-pair and 
two-pair scenarios converged with residuals well below tolerance, confirming the constraint 
set is well-posed. The controlled experiments suggest that increasing coupling density presents challenges for HPIPM's interior-point method in our multi-agent formulation. This motivates TurboADMM's ADMM-based decomposition, which explicitly preserves block-tridiagonal structure at the per-agent level while handling dense coupling through consensus updates.S

\section{Conclusion and Future Work}

We introduced TurboADMM, a specialized single-machine QP solver for multi-agent trajectory optimization that systematically exploits block-diagonal-plus-coupling structure. By synergizing ADMM-based decomposition with Riccati warmstart, qpOASES hotstart mechanism, and shared-memory parallelism, TurboADMM delivers significant speedups over OSQP and MOSEK, enabling real-time multi-agent MPC at 10--204 Hz for 2--14 agents on standard hardware. We release a lightweight C++ library requiring only qpOASES and OpenMP, making TurboADMM practical for both research and production settings.

Promising extensions of this work include support for nonlinear dynamics (via SQP or DDP), embedded deployment on resource-limited platforms, and handling heterogeneous agent dynamics. Also, consider the rising iteration count dilutes the advantage of efficient Riccati initialization, we aim to investigate adaptive penalty scheduling or momentum-based ADMM variants. Reducing the outer-loop iteration count allows the computational efficiency of the Riccati-based warmstart to remain the dominant speedup factor, even as the system scales to larger and denser populations.

\section*{Acknowledgments}

This work is independent to the job of the author's affiliated institution. 

\bibliographystyle{plainnat}
\bibliography{references}

\end{document}